\pdfoutput=1

\documentclass[11pt]{article}

\usepackage[final]{acl}

\usepackage{times}
\usepackage{latexsym}

\usepackage{pgfplots}
\pgfplotsset{compat=1.17}
\usepackage{filecontents}

\usepackage[T1]{fontenc}

\usepackage[utf8]{inputenc}

\usepackage{microtype}
\usepackage{subcaption}
\usepackage{inconsolata}

\usepackage{graphicx}

\usepackage{amsmath}
\usepackage{geometry}
\usepackage{algorithm}
\usepackage{algpseudocode}
\usepackage{lipsum}

\usepackage{pgfplots}
\pgfplotsset{compat=1.12}
\usepgfplotslibrary{statistics,colormaps,fillbetween}
\usepackage{tikz}
\usetikzlibrary{patterns}

\usepackage{booktabs}
\usepackage{colortbl}
\usepackage{graphicx}
\usepackage{adjustbox}
\usepackage{array}
\usepackage{xcolor}
\usepackage{amsfonts}

\usepackage{pgfplots}
\usepackage{amsmath}
\pgfplotsset{compat=1.18}

\usepackage{soul}
\usepackage{xcolor}
\usepackage[normalem]{ulem}

\title{Will Annotators Disagree? Identifying Subjectivity\\in Value-Laden Arguments}

\author{Amir Homayounirad, Enrico Liscio, Tong Wang,\\
{\bf Catholijn M. Jonker,} \and {\bf Luciano C. Siebert} \\
Delft University of Technology, the Netherlands \\
\texttt{\{a.homayounirad,e.liscio,t.wang-12,}\\\texttt{c.m.jonker,l.cavalcantesiebert\}@tudelft.nl}}

\pgfplotsset{compat=1.17}

\begin{document}
\maketitle
\begin{abstract}

Aggregating multiple annotations into a single ground truth label may hide valuable insights into annotator disagreement, particularly in tasks where subjectivity plays a crucial role. In this work, we explore methods for identifying subjectivity in recognizing the human values that motivate arguments. We evaluate two main approaches: inferring subjectivity through value prediction vs. directly identifying subjectivity. Our experiments 
show that direct subjectivity identification significantly improves the model performance of flagging subjective arguments. Furthermore, combining contrastive loss with binary cross-entropy loss does not improve performance but reduces the dependency on per-label subjectivity.
Our proposed methods can help identify arguments that individuals may interpret differently, fostering a more nuanced annotation process\footnote{Our code is publicly available at https://github.com/Amir-Homayouni/subjectivity-value}.

\end{abstract}

\section{Introduction}


Human values, spanning concepts such as benevolence and self-determination, are the
motivations that guide our choice and action, and are ordered by importance to form a system of value priorities \cite{Schwartz1994}. Value-laden arguments are statements grounded in our personal values, which we use to motivate our choices \cite{Bench-Capon2003}. The identification of the values that support our arguments can reveal our deepest motivations, and as such has been recently investigated in the NLP community \cite{kiesel-etal-2023-semeval,kiesel:2024d}.

Supervised NLP methods have been proposed to identify the values that support a text segment \cite{kiesel2022identifying,liscio2022cross}. 
Typically, the ground truth labels are chosen through majority aggregation of the annotations \cite{hoover2020moral,kiesel-etal-2023-semeval}
or the annotators engage in discussions to reach an agreement on the annotation \cite{liscio2021axies,lei-etal-2024-emona}. 
However, due to the subjective nature of valuing \cite{mackie1988subjectivity, stroud1988study}, disagreement in the interpretation of the values that support an argument is natural. 
%
For example, consider the following argument in favor of a multi-party political system:  ``(it) would bring many new and fresh ideas into the forefront''. Alice may associate the argument with the value of universalism since a multi-party system can provide all people with equal opportunities. Bob, instead, may connect it to the values of achievement and personal security, because a multi-party system can more effectively address issues than a single-party system and thus provide more security to its citizens\footnote{The argument and the two annotations are sourced from the ValueEval'23 Shared Task \cite{kiesel-etal-2023-semeval}.}.

Approaches relying on consensus or majority aggregation might 
obscure the inherent subjectivity of the identification of the values behind arguments, leading to misinterpretation or, at worst, promotion of biases stemming from the annotation process.
Identifying subjectivity in value-laden arguments thus has broader implications, particularly in contexts that demand participatory deliberation and collective decision-making.

The assumption that annotations should be aggregated into a unique label is being questioned within the NLP community \cite{Rttger2021TwoCD,weerasooriya-etal-2023-disagreement,van-der-meer-etal-2024-annotator}, in applications ranging from hate speech detection \cite{Koco2021LearningPH,davani-etal-2022-dealing} to sentiment and emotion detection \cite{deng-etal-2023-annotate}.
In line with these works, in this paper we propose methods to identify subjective value-laden arguments---that is, detect arguments where annotators may have different interpretations of the values that support them.

We envision two primary applications of this work.
(1) Identifying arguments that might be subjective can support a more nuanced annotation process by prompting the collection of additional annotations. 
(2) Identifying the arguments that might lead to misinterpretation and divergent views during participatory deliberations can prompt moderators to ask additional questions that could promote self or inter-participant reflection.


\paragraph{Contribution}

We propose two different approaches to detecting subjectivity in value identification within discourse: three methods for inferring subjectivity from individual value annotations, and three methods for directly identifying subjectivity,
which we further enhance with contrastive learning strategies.
We validate the approaches on the Touché23-ValueEval dataset \cite{kiesel-etal-2023-semeval}. Our results show that we can identify subjectivity in value-laden arguments, and that directly identifying subjectivity—rather than inferring it through value prediction—greatly improves performance. Additionally, leveraging contrastive loss does not improve subjectivity prediction performance but brings other advantages.

\section{Related works}

We review related works on identifying values in text and on subjectivity in NLP applications.

\subsection{Identifying Values in Text}

Identifying the value(s) that support a natural language statement has been approached through word count and sentence embedding similarity to dictionaries of value-laden words \cite{araque2020moralstrength}. More recent approaches employ supervised machine learning on annotated datasets \cite{alshomary-etal-2022-moral, Huang2022LearningTA, park-etal-2024-morality,liscio2025value, senthilkumar2024leveraging}.
In particular, \citet{kiesel2022identifying} focuses on identifying and classifying the values underlying arguments.
They successfully fine-tuned a BERT model \cite{Devlin2019BERTPO} on multi-cultural arguments, 
and later extended the dataset for the ValueEval challenge at SemEval'23 \cite{kiesel-etal-2023-semeval}.

In this study, unlike the previous work that uses golden ground truths to identify values, we utilize a dataset consisting of annotation of values to arguments to explore the subjectivity of annotations in recognizing the values behind arguments.

\subsection{Identifying Subjectivity}



Subjectivity is playing an increasingly central role in various NLP tasks \cite{plank-2022}.
Datasets that report individual-level annotation (e.g., \cite{Aroyo2023DICESDD}) facilitate the modeling of individual and group annotation behavior, with different annotation paradigms shown to have a great impact on data quality and model performance \citet{Rttger2021TwoCD}. 
This information allows to represent ground truth as a label distribution, preserving diverse human judgments and minority opinions \citet{weerasooriya-etal-2023-disagreement}.
Different methods account for subjectivity by combining annotator and annotation embeddings \cite{deng-etal-2023-annotate}, modeling multi-annotator architectures \cite{davani-etal-2022-dealing}, and capturing annotators' perspectives by combining their demographic information and their opinions on online content \cite{fleisig-etal-2023-majority}. 
Demographic information of annotators has been employed as a feature, however with mixed results \cite{goyal2022toxicity,Wan2023,orlikowski-etal-2023}.
Subjectivity has also been explored as part of the sampling strategy of Active Learning methods, e.g. to select the next sample to be annotated \cite{baumler-etal-2023-examples,wang2023actor} or the next annotator that should annotate it \cite{van-der-meer-etal-2024-annotator}.




These approaches primarily focus on predicting the opinions of individuals or groups of annotators. Instead, our methods identify whether we can expect disagreement among annotators identifying the values that support an argument, thus not being reliant on specific individuals or groups.

\subsection{Measuring disagreement on moral ambiguity}

In morally ambiguous scenarios, disagreements naturally arise due to the inherent diversity and pluralism of values people hold. Grounding theories like value pluralism suggest that such disagreements reflect legitimate differences in moral frameworks rather than mere errors \cite{kekes1996morality}. Among others, the Moral Foundations Theory illustrates how different moral intuitions (care, fairness, authority, loyalty, sanctity) lead to divergent judgments among groups such as liberals and conservatives, making some conflicts particularly intractable \cite{haidt2007morality,graham2013moral}. Social psychologists further emphasize cognitive biases, such as naïve realism, where individuals perceive their views as objectively correct and opposing views as biased or misinformed, intensifying disagreements \cite{ross2013naive, lord1979biased}. Empirical methods from social sciences have operationalized these disagreements through inter-coder reliability metrics (e.g., Cohen’s kappa, Krippendorff’s alpha), recognizing that systematic variance among annotators can reflect meaningful differences in interpretation rather than random error \cite{krippendorff2004reliability}. Theories of deliberative democracy, notably by \citet{gutmann2004deliberative, rawls2002john}, provide frameworks for managing moral disagreements through mutual respect and reason-giving rather than forcing consensus.

\section{Methods}

We propose two approaches to identify subjective value-laden arguments---that is, detect whether we expect annotators to disagree on the value annotation of a piece of text.
%
In the first approach, Inferred Subjectivity identification (IS, Section~\ref{sec:method:ISV}), we train models to predict value labels for individual annotators and infer subjectivity from the variations in labels predicted across annotators. In the second approach, Direct Subjectivity identification (DS, Section~\ref{sec:method:DS}), we train models to directly classify whether a given argument is subjective.

\begin{figure*}[!htb]
  \includegraphics[width=1\linewidth]{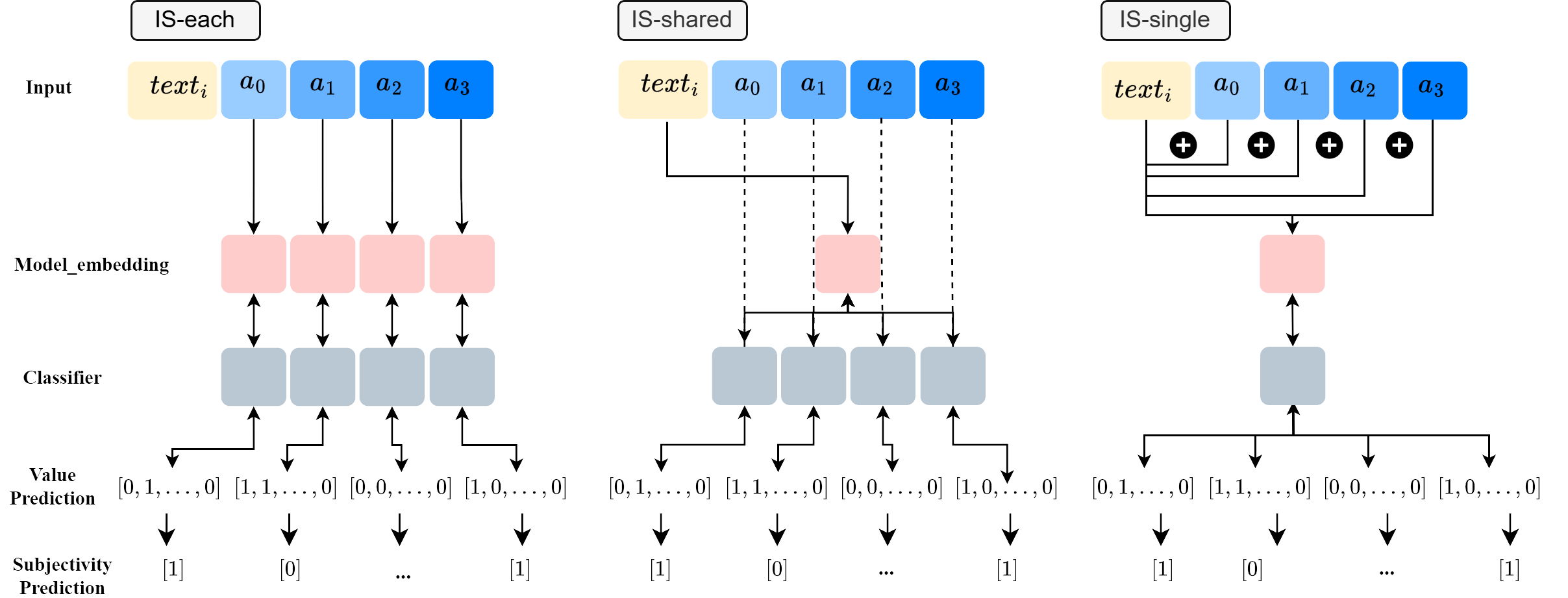}
  \caption{Representation of how the method processes input text to predict values for individual annotators and then subsequently infer subjectivity for each value label}
  \label{fig:subjectivity1}
\end{figure*}

\subsection{Task Formalization}

Consider a dataset $D$ composed of annotated triples $(x_i, y_{ij}, a_j)$, where $x_i$ is a piece of text containing an argument, and $y_{ij}$ is the annotation of annotator $a_j $. Annotators can assign multiple values to each text $x_i$, chosen from a list of values $v_k \in V$. We aim to create a model $f(x) \in (0,1)$ that predicts whether we expect disagreement (0) or agreement (1) among the annotators when annotating the value(s) that support an argument. We consider an argument to be subjective if at least one annotator from a group assigns a different set of values to that text compared to others. 

\subsection{Inferred Subjectivity Identification (IS)}
\label{sec:method:ISV}

In this approach, we explore three multi-annotator architectures to assess subjective value prediction, as displayed in Figure~\ref{fig:subjectivity1}. They all pass the input arguments into sentence embedding and classification heads in different ways to predict multi-label values for each annotator. Appendix~\ref{Results on value prediction} presents the results of each method for value prediction. We then classify $x_i$ as subjective for each $v_k$ where individual annotators' predictions differ and report the subjectivity classification results.

\subsubsection{IS-each: A Dedicated Model for Each Annotator}
\label{sec:method:DBV}

We train a multi-label classification model for each annotator $a_j$ to predict value values based on the annotations they provided.
%
%
Although straightforward, this approach is computationally expensive as it requires training a separate model for each annotator.

\subsubsection{IS-shared: a Shared Model with a Dedicated Head for Each Annotator}

We train a single model for all annotators, thus with shared embeddings but a different multi-label classification head for each annotator. This approach reduces computational complexity by sharing common embeddings across all annotators.


\subsubsection{IS-single: One Model for All Annotators}

In this approach, we train a single model for all annotators, incorporating a unique annotator identifier as part of the input. The input to the model is modified by concatenating the annotator ID  ($a_j$) with the textual data. This method may address some of the computational constraints associated with the previous methods, however, it may not capture annotator nuances by just using annotator ID as input and their annotation as output. 

\subsection{Direct Subjectivity Identification (DS)}
\label{sec:method:DS}

\begin{figure*}[!htb]
  \includegraphics[width=1\linewidth]{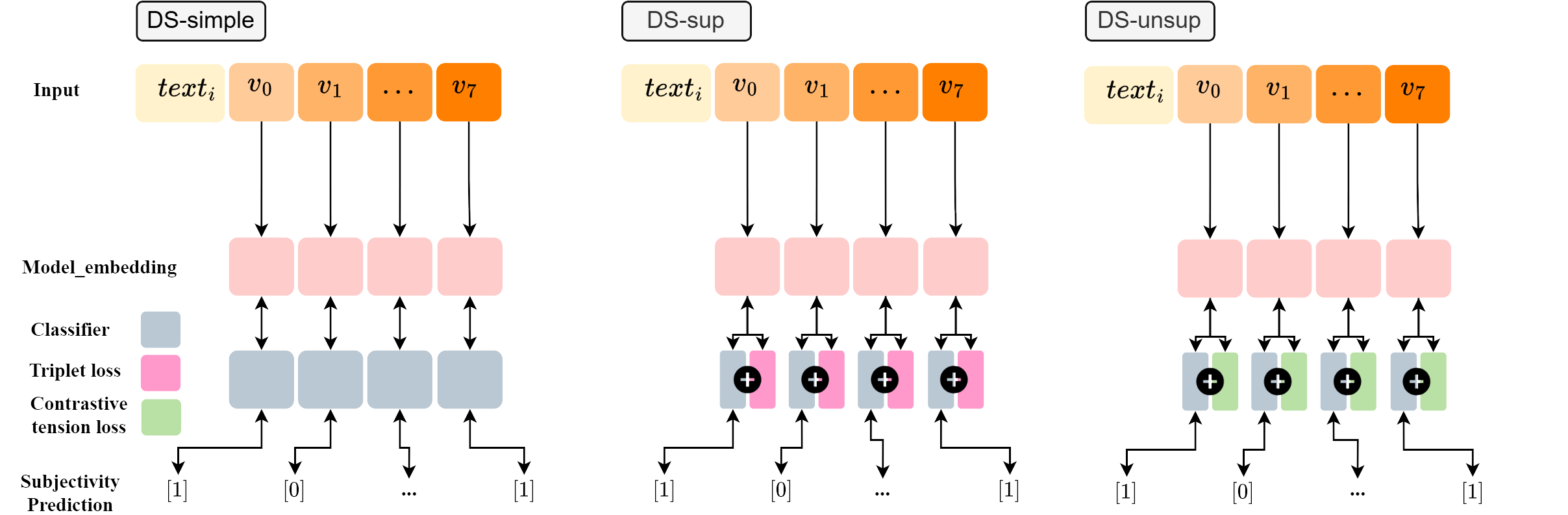}
  \caption{Representation of how the method processes input text to identify subjectivity for each value label}
  \label{fig:subjectivity2}
\end{figure*}

DS directly trains the model to assess whether we expect annotators to disagree in the annotation of $ v_k $ in text $x_i$, as displayed in Figure~\ref{fig:subjectivity2}. We frame the task as a binary classification problem and compare three methods.
In the first, we employ a dedicated model for each value label to predict subjectivity. In the next two, we insert a contrastive learning objective in the model, in a supervised and unsupervised manner, respectively.

\subsubsection{DS-simple: A Model to Predict Subjectivity for Each Value Label}

Analogously to section \ref{sec:method:DBV}, we train a model to predict whether we expect the annotation to be subjective, for each value  $( v_k )$, thus resulting in $|V|$ binary prediction models.

\subsubsection{Contrastive Loss Primer}
\label{sec:contrastive-loss}

Cross-entropy loss has been shown to have several shortcomings, such as not explicitly encouraging discriminative learning, leading to poor generalization performance \cite{Liu2016LargeMarginSL,Cao2019LearningID}. Contrastive learning as an auxiliary training objective during fine-tuning has been shown to mitigate these problems \cite{Gunel2020SupervisedCL}, even in an unsupervised manner \cite{kim-etal-2021-self}.
Contrastive learning focuses on enhancing the representation of sentence embeddings by bringing semantically similar examples closer together and separating dissimilar examples \cite{Hadsell2006}. 

For the following two methods, we consider a primary task as training a binary classifier to predict the subjectivity of the input text using Binary Cross-Entropy (BCE) loss, and as an auxiliary task use a Contrastive Learning (CL) loss:
%
%
\begin{equation*}
L = L_{BCE} + \lambda L_{CL}
\end{equation*}
where \( L_{BCE} \) represents the binary cross-entropy loss, \( L_{CL} \) denotes the contrastive learning loss, and $\lambda$ controls the relative importance of the CL loss compared to the BCE loss. The methods differ in how $L_{CL}$ is calculated---in a supervised and unsupervised fashion, respectively.

\subsubsection{DS-sup: DS-simple + Supervised Contrastive Loss}

In this method, we use triplet loss ($L_{triplet}$) \cite{Schroff_2015}, a supervised version of contrastive loss that exploits data labels to refine the embedding space such that examples labeled with the same class (subjective) are pushed closer together and examples with a different class (non-subjective) are pushed further apart.




Every training sample is composed of a triple $(x_A, x_P, x_N)$, where $x_A$ is the anchor sample, $x_P$ (positive sample) is a randomly selected sample with the same label as $x_A$, and $x_N$ (negative sample) is a randomly selected sample with a different label from $x_A$. $L_{triplet}$ is then defined as:
\begin{equation*}
L_{triplet} = \max \left\{ d(z_A, z_P) - d(z_A, z_N) + m, \, 0 \right\}
\end{equation*}
where $z_A$, $z_P$, and $z_N$ are the normalized representations in embedding space of the anchor, positive, and negative inputs, respectively. $m > 0$ is the margin hyperparameter that enforces a minimum separation between positive and negative pairs.







\subsubsection{DS-unsup: DS-simple + Unsupervised Contrastive Loss}

In this method, we employ contrastive tension loss ($L_{tension}$) \cite{Carlsson2021SemanticRW}, an unsupervised contrastive loss. Whereas the previous method employed label information to shape the embedding space, this method instead ensures that semantically similar samples are pushed closer together while dissimilar examples are pushed further apart. $L_{tension}$ is then defined as:
\begin{equation*}
L_{\mathrm{tension}} \;=\;
-\;\frac{1}{N}
\sum_{i=1}^{N}
\log
\Biggl(
  \frac{
    \exp\!\Bigl(\tfrac{\mathrm{sim}\bigl(z_i, z_{i^+}\bigr)}{\tau}\Bigr)
  }{
    \displaystyle
    \sum_{j=1}^{N}
    \exp\!\Bigl(\tfrac{\mathrm{sim}\bigl(z_i, z_{j}\bigr)}{\tau}\Bigr)
  }
\Biggr),
\end{equation*}

where $z_i$ is the embedding of the \emph{anchor} sample, $z_{i^+}$ is the embedding of a \emph{positive} sample , $z_j$ are all samples in the batch (acting as negatives when $j \neq i^+$), $\mathrm{sim}(\cdot,\cdot)$ is a similarity function, $\tau > 0$ is a temperature hyperparameter, and $N$ is the batch size.






\color{black}

\section{Experimental Setup}

We describe the models, dataset, and evaluation metrics for our experiments. 

\subsection{Models}

We test all proposed methods with BERT-base \cite{Devlin2019BERTPO} as model embedding.
Next, we fine-tune the Llama-3.1-8B-Instruct model \cite{touvron2023} for two variants of the two approaches (IS-single and DS-simple), as further elaborated in Section~\ref{sec:results}.
We compare proposed methods against a baseline that randomly predicts subjectivity. While our experiments are only conducted with two models, we propose a model-agnostic approach that is not limited to these models.
Appendix~\ref{Methodological details} details the used hyperparameters.



\color{black}

\subsection{Dataset}

\begin{figure}[h!]
\begin{tikzpicture}
    \begin{axis}[
        ybar,
        bar width=0.2cm,
        width=0.4\textwidth,
        height=0.2\textheight,
        legend style={at={(0.5,-0.5)}, anchor=north, legend columns=-1, font=\tiny},
        symbolic x coords={Ach, Pow, Sec-p, Sec-s, Con, Ben-car, Ben-dep, Uni},
        xtick=data,
        every axis/.append style={font=\scriptsize},
        ylabel={Count},
        ymin=0,
        ymax=3000,
        ytick={0,500,...,3000},
        xticklabel style={rotate=45, anchor=east},
    ]

    \addplot[ybar, fill=blue!30, draw=blue] coordinates {
        (Ach, 2038) (Pow, 2232) (Sec-p, 2231) (Sec-s, 2232) 
        (Con, 2323) (Ben-car, 2417) (Ben-dep, 2516) (Uni, 2271)
    };

    \addplot[ybar, fill=red!30, draw=red] coordinates {
        (Ach, 743) (Pow, 549) (Sec-p, 550) (Sec-s, 549) 
        (Con, 458) (Ben-car, 364) (Ben-dep, 265) (Uni, 510)
    };

    \end{axis}

    \begin{axis}[
        axis y line*=right,
        axis lines=none,  
        ytick=\empty,     
        ymin=0, ymax=0.5,
        width=0.4\textwidth,
        height=0.2\textheight,
        symbolic x coords={Ach, Pow, Sec-p, Sec-s, Con, Ben-car, Ben-dep, Uni},
        xtick=\empty,  
        clip marker paths=true,
        axis on top     
    ]
    
    \addplot[
        red,
        thick,
        mark=square*,
        mark options={fill=red},
        nodes near coords,
        every node near coord/.append style={
            font=\tiny,,
            anchor=south,
            text=red,
            /pgf/number format/fixed
        }
    ] coordinates {
        (Ach, 0.364)
        (Pow, 0.246)
        (Sec-p, 0.247)
        (Sec-s, 0.246)
        (Con, 0.197)
        (Ben-car, 0.151)
        (Ben-dep, 0.105)
        (Uni, 0.225)
    };
    \legend{\tiny Ratio (S/NS)}
    \end{axis}
\end{tikzpicture}
\caption{Counts and ratio of subjective (S, red) and non-subjective (NS, blue) labels per each selected value.}
\label{fig: Subjectivity Distribution}
\end{figure}
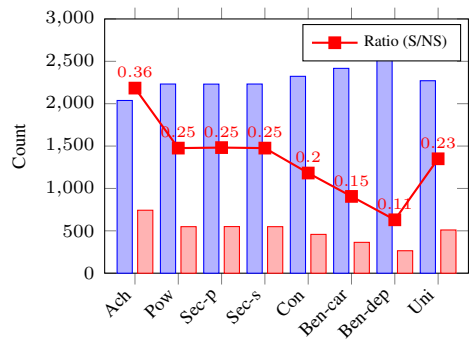

We use the Touché23-ValueEval dataset \cite{Mirzakhmedova2023TheTD}, which, to the best of our knowledge, is the only dataset available that includes value annotations from multiple annotators for each instance. The dataset is composed of 9324 natural language arguments annotated with a taxonomy of 54 values (multi-label annotation)
 derived from the Schwartz Value Survey \cite{Schwartz2012RefiningTT} and distributed in four hierarchical levels. Argument datasets are almost exclusively from a Western background on controversial topics namely religious texts, political discussions, free-text arguments, newspaper editorials, and online democracy platforms. We utilize annotations representing the crowd workers' original annotations (before being aggregated into a single ground truth label) that have all been carried out by annotators from a Western background. 
To effectively demonstrate our methods while minimizing the computational load,
we selected the eight most frequently annotated values from level 2. These values were annotated by the four annotators who, among the 39 annotators, had the highest annotation overlap. The selected values are Achievement (\texttt{Ach}), Power: resources (\texttt{Pow}), Security: personal (\texttt{Sec-p}), Security: societal (\texttt{Sec-s}), Conformity: rules (\texttt{Con}), Benevolence: caring (\texttt{Ben-car}), Benevolence: dependability (\texttt{Ben-dep}),  Universalism: concern (\texttt{Uni}). Figure~\ref{fig: Subjectivity Distribution} presents the distribution between the subjective and non-subjective annotations for each of the selected values, and Appendix \ref{Dataset details} provides more information on the dataset, annotators, and selected values. 

Furthermore, for the DS approaches, we augment the minority class to balance the dataset by paraphrasing the minority class sentences \cite{alisetti2021paraphrase} (see Appendix~\ref{Paraphrasing} for additional details).
Data augmentation is not possible in the IS approach due to the multi-label nature of the task---augmenting data for one label would also impact the other labels annotated on the same sentence.

\subsection{Evaluation Metrics}

We report precision, recall, and $F_1$-score on the test set (which is fixed, as detailed in Appendix~\ref{Dataset details}), per value and averaged over the values.
In addition, we report the Spearman correlation ($\rho$) between the per-value $F_1$-scores and the subjective to non-subjective ratios (Figure~\ref{fig: Subjectivity Distribution}) to investigate the correlation between subjectivity level and performance.

\section{Results}
\label{sec:results}

Table~\ref{table:average-results} reports the performance of the different approaches averaged over the selected values, whereas Table~\ref{table:results} reports the per-value performance.

\begin{table}[h]
\centering
\small
\begin{tabular}{l|c|c|c|c}
\toprule
\textbf{Method} & \textbf{P}& \textbf{R}& \textbf{$F_1$} & \textbf{$\rho$} \\

\midrule

IS-each (BERT) & 0.34 & 0.40 & 0.36 & 0.88  \\
IS-shared (BERT) & 0.31 & 0.22 & 0.25 & 0.85 \\
IS-single (BERT) & 0.35 & 0.30 & 0.32 & 0.86 \\
IS-single (Llama) & 0.40 & 0.26 & 0.30 & 0.88 \\

\midrule

DS-simple (BERT) & 0.85 & 0.61 & 0.70 & -0.60  \\
DS-sup (BERT) & 0.85 & 0.62 & 0.71 & -0.32  \\
DS-unsup (BERT) & 0.70 & 0.61 & 0.65 & 0.24 \\
DS-simple (Llama) & 0.84 & 0.76 & 0.80  & -0.97\\

\midrule

Baseline (random)  & 0.53 & 0.50 & 0.51 & -0.12 \\

\bottomrule

\end{tabular}
\caption{Average precision (P), recall (R), and $F_1$-score across the selected value labels, together with the correlation ($\rho$) between $F_1$-scores and subjectivity ratio.}
\label{table:average-results}
\end{table}

\begin{table*}[tp]
\centering
\small
\begin{tabular}{l|ccc|ccc|ccc|ccc}
\toprule
\textbf{Method} & \multicolumn{3}{c|}{\textbf{Achievement}} & \multicolumn{3}{c|}{\textbf{Power: resources}} & \multicolumn{3}{c|}{\textbf{Security: personal}} & \multicolumn{3}{c}{\textbf{Security: societal}} \\

 & \textbf{P} & \textbf{R} & \textbf{F$_{1}$}  & \textbf{P} & \textbf{R} & \textbf{F$_{1}$}  & \textbf{P} & \textbf{R} & \textbf{F$_{1}$} & \textbf{P} & \textbf{R} & \textbf{F$_{1}$}  \\
 
\midrule

IS-each (BERT) & 0.40 & 0.56 & 0.46 
   & 0.41 & 0.50 & 0.45 
   & 0.35 & 0.44 & 0.39
   & 0.36 & 0.50 & 0.42  \\

IS-shared (BERT) & 0.37 & 0.29 & 0.32 
   & 0.38& 0.26 & 0.31
   & 0.32& 0.20 & 0.24
   & 0.41 & 0.34 & 0.37  \\

IS-single (BERT) & 0.38 & 0.39 & 0.38 
   & 0.40 & 0.33& 0.36
   & 0.37 & 0.40 & 0.38 
   & 0.38 & 0.36 & 0.37  \\

      IS-single (Llama) & 0.41 & 0.39  & 0.40
    & 0.44 & 0.29  & 0.35 
    & 0.46 & 0.41  & 0.33
    & 0.45 & 0.31  & 0.37 \\

\midrule

DS-simple (BERT) & \textbf{0.81} & 0.57 & 0.67 
      & 0.82& \underline{0.61} & \underline{0.69}  
      & \textbf{0.83} & 0.66 & \underline{0.73} 
      & \textbf{0.89} & 0.58 & 0.70 \\

DS-sup (BERT) & \underline{0.78} & \underline{0.62} & \underline{0.69}
    & \underline{0.84} & 0.58 & \underline{0.69}
    & \textbf{0.83} & 0.64 & 0.72 
    & \underline{0.84} & 0.65 & \underline{0.73}  \\

DS-unsup (BERT) & 0.65  & 0.60  & 0.62
    & 0.73& 0.55 & 0.63 
    & 0.65 & \underline{0.72}& 0.68
    & 0.71 & \underline{0.67} & 0.68 \\

DS-simple (Llama)
    & 0.71 & \textbf{0.76} & \textbf{0.74}
    & \textbf{0.94} & \textbf{0.65} & \textbf{0.77}
    & 0.73 & \textbf{0.80} & \textbf{0.76} 
    & 0.82 & \textbf{0.74} & \textbf{0.78}
      \\

\midrule

Baseline (random) & 0.55 & 0.49  & 0.51 
         & 0.53 & 0.50  & 0.52
        & 0.51  & 0.48 & 0.49
        & 0.55 & 0.49 &  0.52 \\

\bottomrule
\toprule

 & \multicolumn{3}{c|}{\textbf{Conformity: 
rules}} & \multicolumn{3}{c|}{\textbf{Benevolence: caring}} & \multicolumn{3}{c|}{\textbf{Benevolence: depend.}} & \multicolumn{3}{c}{\textbf{Universalism: concern}} \\

 & \textbf{P} & \textbf{R} & \textbf{F$_{1}$}  & \textbf{P} & \textbf{R} & \textbf{F$_{1}$}  & \textbf{P} & \textbf{R} & \textbf{F$_{1}$} & \textbf{P} & \textbf{R} & \textbf{F$_{1}$}  \\
 
\midrule

IS-each (BERT) & 0.30 & 0.35& 0.33 
   & 0.33 & 0.42 & 0.37
   & 0.20 & 0.24 & 0.22
   & 0.36 & 0.2 & 0.24  \\

IS-shared (BERT) & 0.28 & 0.18 & 0.22
   & 0.28 & 0.18 & 0.22
   & 0.14& 0.15 & 0.14 
   & 0.32 & 0.16 & 0.20  \\

IS-single (BERT) & 0.30 & 0.26& 0.27
   & 0.35 & 0.27 & 0.30 
   & 0.23& 0.19 & 0.21  
   & 0.37 & 0.20 & 0.25 \\

   IS-single (Llama)  & 0.42 & 0.19  & 0.26 
    & 0.35 & 0.22  & 0.27   
    & 0.33 & 0.13   & 0.18
    & 0.35 & 0.17 & 0.23  \\

\midrule

DS-simple (BERT) & \textbf{0.89} & 0.60 & 0.71
      & 0.86 & \underline{0.62} & \underline{0.71} 
      & \textbf{0.90} & 0.60 & 0.72 
      & \underline{0.85} & \underline{0.59} & \underline{0.70}  \\

DS-sup (BERT) & \textbf{0.89} & 0.61 & \underline{0.72}  
    & \textbf{0.89} & 0.56 & 0.69
    & 0.86 & \underline{0.67} & \underline{0.75}
    & \underline{0.85} & \underline{0.59} & 0.69 \\

DS-unsup (BERT) & 0.70 & \underline{0.68}  & 0.69
         & 0.78 & 0.52  & 0.62
        & 0.66  & 0.55 & 0.60
        & 0.70 & \underline{0.59} &  0.64 \\

DS-simple (Llama)  
    & 0.88 & \textbf{0.75} & \textbf{0.81}
    & \underline{0.88} & \textbf{0.78} & \textbf{0.83}
    & \underline{0.88} & \textbf{0.86} & \textbf{0.87}
    & \textbf{0.91} & \textbf{0.71} & \textbf{0.80}
    
      \\

\midrule

Baseline (random) & 0.50  & 0.50  & 0.50
         & 0.52  & 0.50  & 0.51
        & 0.52  & 0.52 & 0.52
        & 0.53  & 0.50 &  0.52\\

\bottomrule
\end{tabular}%
\caption{Precision (P), recall (R), and $F_1$-score of the subjectivity prediction per value. In bold, highlight the best-performing method and underline the second-best-performing method for each value and metric.}
\label{table:results}
\end{table*}

\subsection{Comparison across Methods}
First, we compare the two primary approaches---inferring subjectivity from value prediction (IS) versus direct subjectivity identification (DS). We notice that the latter demonstrates superior performance and that even the random baseline mostly outperforms the IS approach (except for IS-each for recall in Ach, Pow, and Sec-s). This may indicate that predicting value solely through annotation may not be sufficient to capture individual subjective preferences and infer subjectivity. However, the DS approach consistently outperforms the baseline in all metrics.

Second, we compare the methods that infer subjectivity from value prediction (IS). We observe that IS-each consistently outperforms the others. This is presumably because having a dedicated model for each annotator better captures their annotation tendencies when compared to having a shared embedding layer (IS-shared) or having a fully shared model (IS-single). Moreover, IS-single consistently outperforms IS-shared, which shows that differentiating the input text with an annotator ID shows a better performance compared to having a dedicated head for each annotator. 

Finally, we compare the approaches that directly infer subjectivity (DS). We notice that DS-simple and DS-sup generally perform similarly and better than DS-unsup. However, we observe that for more subjective values (such as Ach) the performances with the three methods are comparable, whereas for less subjective values (such as Ben-dep) the results of DS-unsup are significantly worse. This is also supported by the correlation scores in Table~\ref{table:average-results}, which show a moderate positive correlation between the DS-unsup results and per-value subjectivity.



\subsection{Comparison across Values}

Next, we compare the results across values.
Table~\ref{table:average-results} shows that the performances with the IS approach are consistently correlated with the subjectivity of the value annotations. With the DS approach, instead, only the DS-unsup results are correlated with subjectivity---that is, for the best-performing methods (DS-simple and DS-sup), lower subjectivity leads to better performances.

We conjecture that the correlation is strongly positive for all IS methods because a more balanced class distribution leads to better value prediction, which in turn leads to better subjectivity identification. Instead, for the DS methods, which directly identify subjectivity, we observe that precision is consistently higher than recall, exceptions are DS-simple (Llama) for \texttt{Ach} and DS-unsup, DS-simple (Llama) for \texttt{Sec-p}. This is likely negatively correlated with subjectivity because the model may overfit to simpler examples hurting performance on more subtle (high subjective) values.


Finally, DS-simple has a stronger negative correlation with per-value subjectivity than DS-sup, suggesting it systematically achieves a higher $F_1$-score on values for which fewer annotators disagree. On the other hand, DS-sup is more balanced across value subjectivity. We conjecture this is due to the fact that supervised contrastive loss pushes the embeddings for the same label closer together, regardless of the value's overall subjective level. 
This shapes the embedding space more consistently across both highly subjective and less subjective values, resulting in a mitigation of overfitting.

\subsection{Comparison across Models}
Finally, we compare the results across models.
For the IS approach, we decided to test the Llama model with the IS-single method to investigate whether using a more powerful model can compensate for the difference between using a single model for all annotators (IS-single) or one model per annotator (IS-each). However, we observe no improvement over the IS-single results with the BERT model. We conjecture that this is due to the structure of IS-single---that is, adding an annotator ID to each argument to differentiate between annotators. Such information is evidently not sufficient to differentiate across annotators. 

Next, given the comparable performances between DS-simple and DS-sup, we decide to train Llama with DS-simple for simplicity. Differently from the IS approach, in the DS approach, we observe a performance improvement over BERT, particularly in recall across all values and in precision for values \texttt{Pow}, and \texttt{Uni}, in line with the difference in state-of-the-art between the two models.

    

\color{black}

\section{Discussion}

In this work, we explored two distinct approaches to address the challenge of identifying subjectivity in value-laden arguments. We discuss our results across methods and value labels. 


\subsection{Comparison across Methods}

\paragraph{DS is better, but IS also has its merits.} The superior performance of the DS approach can be likely attributed to its focused objective, which simplifies the model's learning process by concentrating on distinguishing between subjective and non-subjective instances. This targeted focus allows the model to capture patterns indicative of subjectivity, leading to improved recall and $F_1$-score rates. While it is intuitive that direct subjectivity identification might outperform inference-based methods, we argue that inferring subjectivity from value predictions remains valuable in scenarios where understanding individual value preferences provides meaningful insights. For instance, exploring why an argument is subjective by looking at each annotator's value prediction and understanding their different interpretation. However, improving subjectivity identification through value prediction may require the development of methodologies that explicitly model the relationship between value prediction and subjectivity identification. For example, incorporating individual subjective preferences derived not only from value annotations but also from additional factors such as annotators' backgrounds can enhance this link.

Moreover, the subjectivity of value-laden arguments may vary across different contexts and demographic groups \cite{liscio2021axies}. The second approach, which directly predicts subjectivity irrespective of individual annotators, offers a pathway for active learning strategies to update and fine-tune the model for new contexts and diverse populations.


\begin{figure}[!tb]
  \centering
  \begin{subfigure}[b]{\linewidth}
  \centering
    \includegraphics[width=0.8\linewidth]{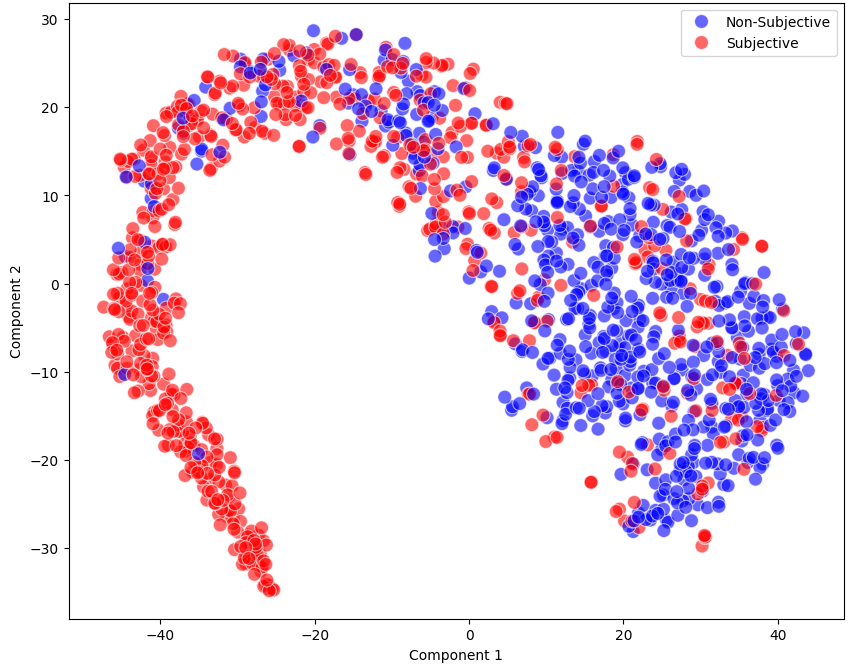}
    \caption{BERT-base}
    \label{fig:bert-base}
  \end{subfigure}
  \vfill
  \begin{subfigure}[b]{\linewidth}
  \centering
    \includegraphics[width=0.8\linewidth]{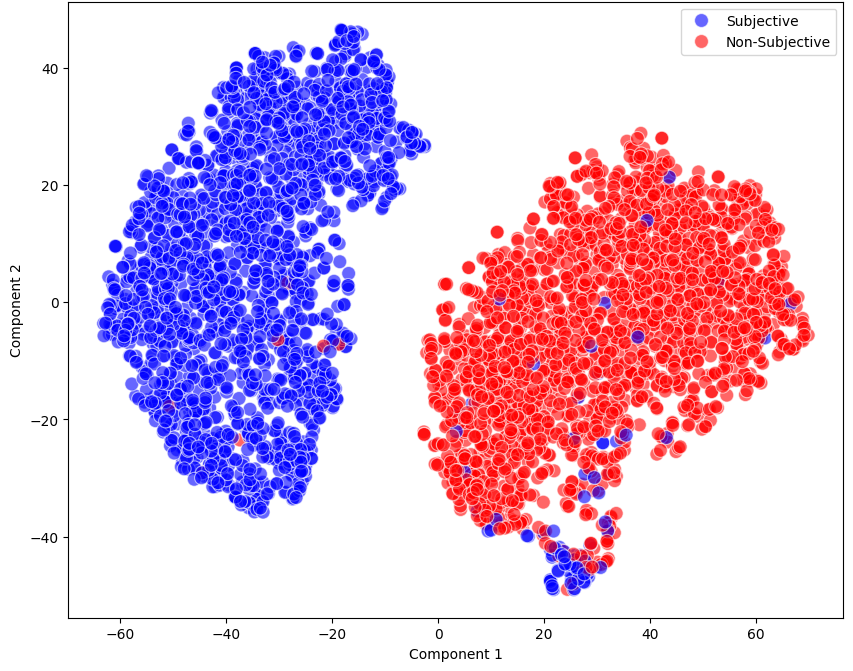}
    \caption{BERT embeddings trained with DS-sup.}
    \label{fig:bert-base-dbcssc}
  \end{subfigure}
  \vfill
  \begin{subfigure}[b]{\linewidth}
  \centering
    \includegraphics[width=0.8\linewidth]{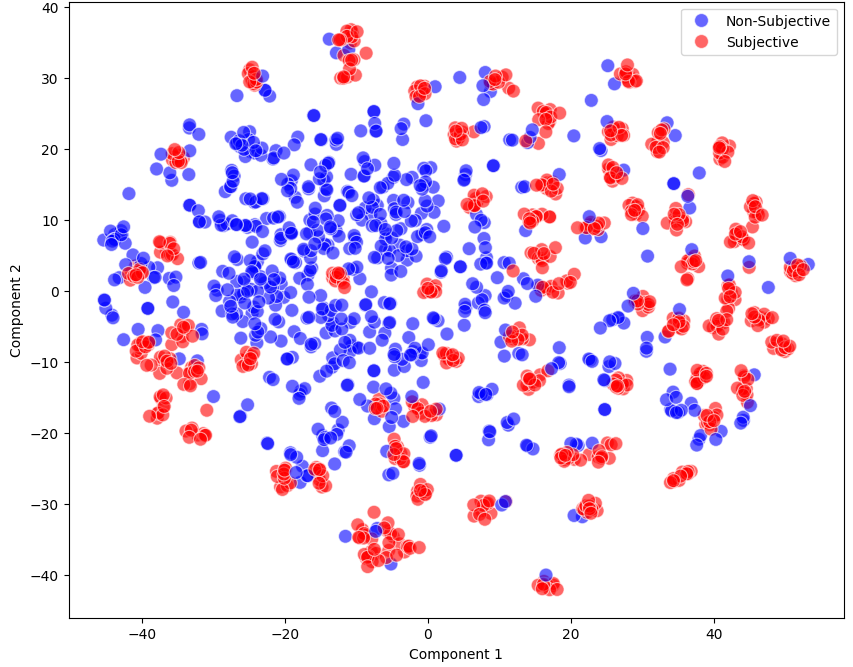}  
    \caption{BERT embeddings trained with DS-unsup.}
    \label{fig:bert-base-dbcsuc}
  \end{subfigure}
  \caption{Comparison of 2D sentence embeddings visualizations obtained through t-SNE. (Top) Embeddings from the original BERT. (Middle) Embeddings from the BERT instance fine-tuned with DS-sup. (Bottom) Embeddings from the BERT instance fine-tuned with DS-unsup. Red dots correspond to subjective and blue to non-subjective.}
  \label{fig:combined}
\end{figure}

\paragraph{Contrastive loss brings additional advantages.}
 Despite the comparable subjectivity prediction performance, optimizing binary cross-entropy loss with a contrastive learning objective makes the resulting embeddings more suitable for calculating similarities between samples. We confirm this by visualizing the BERT embedding space for value \texttt{Ben-dep} resulting from the three DS methods, in Figure~\ref{fig:combined} (similar patterns were observed for all considered values). DS-sup pushes the embeddings to have a better separation between subjective and non-subjective classes, while DS-unsup groups semantically similar samples.


Improved embeddings can be leveraged to support the annotation process during an active learning procedure.
For instance, we can use the DS-sup embeddings to present annotators with instances that lie near the decision boundary between subjective and non-subjective classes, enabling annotators to reflect on samples that highlight areas of uncertainty. Additionally, utilizing the DS-unsup to retrieve semantically similar samples that are identified differently in terms of subjectivity, provides annotators with contextualized reference points. This approach may support a more nuanced annotation process by allowing reflection on samples where human judgment is most needed.
Moreover, this strategy can be extended beyond annotation to support participatory democracy, equipping deliberation moderators with useful information to promote reflection and facilitate more thoughtful discussions among participants. 

\subsection{Comparison across Values}

Our findings are supported by \citet{Schwartz1994}, who asserts that values like Achievement and Power are more closely tied to personal interests, whereas Universalism and Conformity are associated with broader societal concerns and the welfare of others. Security and Conformity are boundary values. They are primarily concerned with others’ interests, but their goals also regulate the pursuit of their own interests. Hence, due to the individualized nature of some values which are tied to personal experiences and individual goals such as Achievement tend to exhibit higher subjectivity compared to Universalism which is grounded in broader ethical principles that are more widely shared across different cultures and societies.

\section{Conclusion}

We introduce multiple approaches to identify subjectivity in value-laden arguments. Applying our methods to the Touché23-ValueEval dataset, we demonstrated that directly identifying subjectivity, as opposed to inferring it through value prediction, significantly enhances performance. Implementing a dual-task strategy that combines contrastive loss with BCE loss does not directly improve subjectivity identification, but leads to a model that is less dependent on per-label subjectivity. Finally, using a state-of-the-art model improves performance for direct subjectivity identification, but not for inferred subjectivity identification.


We envision a combination of our proposed approaches as future work, e.g. by combining value and subjectivity prediction to capture individual annotators' perspectives in combination with their annotations, thereby potentially improving the identification of subjectivity in value-laden arguments. Datasets encompassing diverse perspectives, including varying demographics, lived experiences, and moral values \cite{waseem-2016, Patton2019}, such as the recently introduced D3CODE dataset \cite{davani2024}, facilitate the integration of these two aspects.

Finally, we acknowledge that differences in annotation (i.e., subjectivity) can be confounded with noise in the annotation process. Previous work addressed this issue by facilitating deliberation among crowd workers \cite{Schaekermann2018} or by assessing the validity of the explanations provided by annotators for their responses \cite{weber-genzel-etal-2024}. Our work can be instrumental in identifying potentially subjective annotations to support such approaches.


\section{Limitations}

Our evaluation is confined to the Touché23-ValueEval dataset \cite{Mirzakhmedova2023TheTD}, primarily due to the scarcity of datasets within the value community that include annotator-level annotations. This limitation underscores the importance of developing datasets that collect a diverse set of annotators' value annotations. Such datasets would facilitate a more robust evaluation of methods aimed at modeling subjectivity.

The first two methods from ISV including —ISV-each, ISV-shared, and DS-simple are computationally expensive due to their reliance on dedicated embeddings and classifiers. Although these methods provide insights into annotator-specific tendencies, their practicality in large-scale applications is limited. Future research could focus on optimizing these models for scalability or developing lightweight alternatives. One approach for optimizing ISV is to merge annotators who share the same perspective into similar groups and apply an active learning strategy to find the most diverse and useful information to infer subjectivity. The same approach can also be applied for DS-simple to only train the data with the most informative information in terms of subjectivity detection, as Ds-unsup which semantically improves embedding can be utilized for this.

Our study focuses exclusively on classification tasks, which limits the generalizability of our findings to other NLP tasks, such as summarization and question-answering, and to fields beyond NLP, such as Reinforcement Learning and Inverse Reinforcement Learning. Investigating how subjectivity manifests in these tasks and domains remains an important avenue for future work

To use the best of our proposed methods to identify the sources of disagreement and promote reflection, our annotators must be diverse and represent different ranges of moral values, beliefs, and backgrounds. This diversity should align with the perspectives of individuals involved in the deliberation process, ensuring that annotators reflect a broad spectrum of participants. The dataset we use in this study has no intention to be used for a specific deliberation setting, and we had no control over the notion of alignment between annotators' subjective viewpoints and participants in deliberation. Hence, as values are context-specific \citet{Vargo2015, Horbel2016, Chandler2011, Edvardsson2010, liscio2021axies}, to be able to identify subjective value-laden arguments more meaningful with respect to the new deliberation setting, we encourage users of this method to either train the methods in a new context or fine-tune the model with representative annotators of deliberation.

\section*{Acknowledgments}

This project is part of the AiBLE lab, which receives support from the TU Delft AI Labs program. 
This work was partly funded by AlgoSoc, ESDiT, and the Hybrid Intelligence Centre (project numbers 024.005.017, 024.004.031, and 024.004.022). Any opinions, findings, and conclusions or recommendations expressed in this material are those of the author(s) and do not necessarily reflect the views of OCW or those of the AlgoSoc consortium as a whole.
GitHub Copilot was used as an assistant programming tool.

\clearpage

\appendix

\renewcommand{\thefigure}{A\arabic{figure}}
\setcounter{figure}{0}
\renewcommand{\thetable}{A\arabic{table}}
\setcounter{table}{0}

\section{Additional Experimental Details}

\subsection{Dataset details}
\label{Dataset details}

we use The Touché23-ValueEval dataset by\citep{Mirzakhmedova2023TheTD},  which comprises 5270 natural language arguments that are annotated for consolidated taxonomy of 54 values (multi-class annotation). Value taxonomy is categorized on the more abstract levels 2–4  which are derived mainly from the Schwartz Value Survey \cite{Schwartz2012RefiningTT}. Table \ref{tab:data_structure} shows the example of the original dataset. annotations have all been carried out by annotators from a Western background. This dataset is distributed under CC BY-SA 4.0. The data is split so that approximately 78\% of the samples are used for training, 22\% for testing, and within the training set, a further 10\% is reserved for validation. The annotation process involved crowdsourcing on MTurk using a custom three-part interface designed for speed and expertise. The interface presented arguments in a scenario and asked annotators to identify relevant values by answering a yes/no question. Annotators were instructed to select one to five values per argument. The average time spent by annotators on each argument was 2 minutes and 40 seconds. For additional information regarding the dataset, please refer to \cite{kiesel-etal-2023-semeval}

We also use the premise, and we reference it as the argument. 

\begin{table}[h!]
\centering
\scalebox{0.7}{ 
\begin{tabular}{|l|l|p{3cm}|p{3cm}|}
\hline
\textbf{\scriptsize Argument ID} & \textbf{\scriptsize Worker ID} & \textbf{\scriptsize Premise} & \textbf{\scriptsize Simplified\_Value\_lvl2\_ann} \\ \hline
A01001 & W014 & if entrapment can serve to more easily capture... & {[}0, 0, 0, 0, 1, 0, 0, 0, 0, 0, 0, 0, 0, 0, 0, ...{]} \\ \hline
A01001 & W020 & if entrapment can serve to more easily capture... & {[}0, 0, 0, 0, 1, 1, 0, 0, 0, 0, 0, 1, 1, 0, 0, ...{]} \\ \hline
A01001 & W024 & if entrapment can serve to more easily capture... & {[}0, 0, 0, 0, 1, 0, 0, 0, 0, 0, 0, 0, 0, 0, 0, ...{]} \\ \hline
A01002 & W014 & we should ban human cloning as it will only ca... & {[}0, 0, 0, 0, 1, 0, 0, 0, 0, 0, 0, 0, 0, 0, 0, ...{]} \\ \hline
A01002 & W020 & we should ban human cloning as it will only ca... & {[}0, 0, 0, 0, 0, 0, 0, 0, 0, 0, 0, 0, 0, 0, 0, ...{]} \\ \hline
... & ... & ... & ... \\ \hline
\end{tabular}
}
\caption{Data structure of argument annotation}
\label{tab:data_structure}
\end{table}

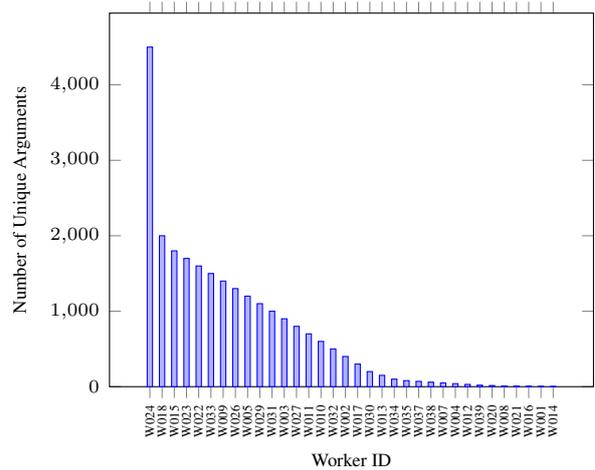
\begin{figure}[h!]
\centering
\begin{tikzpicture}
    \begin{axis}[
        ybar,
        width=0.4\textwidth, 
        height=0.2\textheight, 
        symbolic x coords={W024,W018,W015,W023,W022,W033,W009,W026,W005,W029,W031,W003,W027,W011,W010,W032,W002,W017,W030,W013,W034,W035,W037,W038,W007,W004,W012,W039,W020,W008,W021,W016,W001,W014},
        xtick=data,
        x tick label style={rotate=90, anchor=east, font=\scriptsize, scale=0.7},
        xlabel={\scriptsize Worker ID},
        ylabel={\scriptsize Number of Unique Arguments},
        ylabel style={font=\scriptsize},
        yticklabel style={font=\scriptsize},
        bar width=2pt,
        ymin=0,
        scale only axis
    ]
    \addplot table [x=WorkerID, y=UniqueArguments, col sep=comma] {data.csv};
    \end{axis}
\end{tikzpicture}
\caption{Number of Unique Arguments Annotated by Each Worker}
\label{fig:worker_annotations}
\end{figure}

Following Figure \ref{fig:worker_annotations} we selected the first four annotators who annotated the most. 

\begin{figure}[h!]
\centering
\begin{tikzpicture}
    \begin{axis}[
        width=0.4\textwidth,
        height=0.2\textheight,
        xlabel={\scriptsize Value categories},
        ylabel={\scriptsize Count},
        symbolic x coords={0,1,2,3,4,5,6,7,8,9,10,11,12,13,14,15,16,17,18,19},
        xtick=data,
        bar width= 3 pt, 
        enlarge x limits=0.1,
        yticklabel style={font=\scriptsize},
        xticklabel style={font=\scriptsize},
        title={\scriptsize Value Category Distribution by Worker ID},
        legend style={
            at={(0.5,-0.4)}, 
            anchor=north, 
            legend columns=2, 
            font=\tiny, 
            nodes={scale=1, transform shape} 
        },
        legend cell align={left}
    ]
    \addplot+[ybar, fill=blue!60, mark=*, mark size=1pt] coordinates {(0,112) (1,66) (2,70) (3,5) (4,244) (5,84) (6,272) (7,13) (8,222) (9,267) (10,63) (11,192) (12,41) (13,42) (14,456) (15,214) (16,126) (17,105) (18,49) (19,45)};
    \addplot+[ybar, fill=red!60, mark=square*, mark size=1pt] coordinates {(0,190) (1,163) (2,267) (3,46) (4,557) (5,281) (6,349) (7,94) (8,782) (9,383) (10,103) (11,434) (12,220) (13,82) (14,662) (15,455) (16,250) (17,214) (18,91) (19,199)};
    \addplot+[ybar, fill=green!60, mark=triangle*, mark size=1pt] coordinates {(0,149) (1,223) (2,231) (3,9) (4,488) (5,102) (6,278) (7,63) (8,522) (9,275) (10,144) (11,362) (12,156) (13,95) (14,470) (15,471) (16,272) (17,125) (18,58) (19,138)};
    \addplot+[ybar, fill=orange!60, mark=star, mark size=1pt] coordinates {(0,711) (1,632) (2,840) (3,101) (4,1659) (5,631) (6,1061) (7,467) (8,1529) (9,957) (10,497) (11,1234) (12,449) (13,333) (14,2025) (15,1452) (16,1021) (17,508) (18,382) (19,625)};
    
    \legend{Worker W015, Worker W018, Worker W023, Worker W024}
    \end{axis}
\end{tikzpicture}
\caption{Value Category Distribution by Worker ID}
\label{fig:value_category_desitribution}
\end{figure}
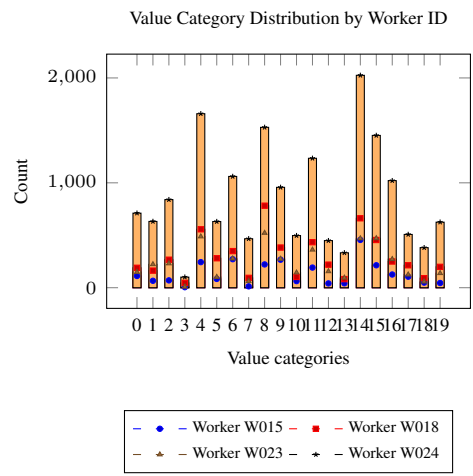

Based on Figure \ref{fig:value_category_desitribution} we selected top 8 Most Annotated Value Categories, Category 14: 3613 annotations : Benevolence: caring(Value\_5) , Category 8: 3055 annotations : Security: personal(Value\_2), Category 4: 2948 annotations: Achievement(Value\_0), Category 15: 2592 annotations : Benevolence: dependability(Value\_6), Category 11: 2222 annotations : Conformity: rules(Value\_4), Category 6: 1960 annotations : Power: resources(Value\_1), Category 9: 1882 annotations : Security: societal(Value\_3), Category 16: 1669 annotations : Universalism: concern (Value\_7)

\subsubsection{Fleiss Kappa score}
\label{Fleiss_Kappa_score}

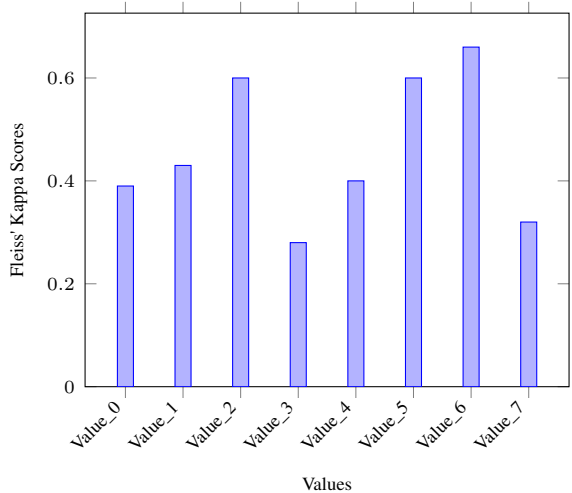
\begin{figure}[h!]
\centering
\begin{tikzpicture}
    \begin{axis}[
        ybar,
        width=0.4\textwidth,
        height=0.2\textheight,
        symbolic x coords={Value\_0,Value\_1,Value\_2,Value\_3,Value\_4,Value\_5,Value\_6,Value\_7},
        xtick=data,
        every axis/.append style={font=\scriptsize},
        x tick label style={rotate=45, anchor=east, font=\scriptsize, scale=1},
        xlabel={\scriptsize Values},
        ylabel={\scriptsize Fleiss\textquotesingle{} Kappa Scores},
        ylabel style={font=\scriptsize},
        yticklabel style={font=\scriptsize},
        bar width=6pt,
        ymin=0,
        scale only axis,
    ]
    \addplot table [x=Values, y=FleissKappa, col sep=comma] {kappa_data.csv};
    \end{axis}
\end{tikzpicture}
\caption{Fleiss\textquotesingle{} Kappa Scores per values}
\label{fig:fleiss_kappa_barplot}
\end{figure}

Fleiss' kappa is a statistical metric used to evaluate the consistency of agreement among multiple raters when they assign categorical ratings to various items \cite{Fleiss1971}. As can be seen \ref{fig:fleiss_kappa_barplot} Value\_0, Value\_7, Value\_3 has fair agreement, and Value\_1, Value\_2, Value\_5 has a moderate agreement and Value\_6 has a substantial agreement.

\subsection{Paraphrasing}
\label{Paraphrasing}

We utilize the \citet{Vamsi_T5_Paraphrase_Paws} model, a T5-based transformer fine-tuned on the PAWS dataset for paraphrase generation. This model generates diverse paraphrased versions of input sentences \ref{table:dbcv-hyperparams} list the hyperparameter used.

\begin{table}[h!]
\centering
\begin{tabular}{|l|l|}
\hline
\textbf{Hyperparameter} & \textbf{Value} \\ \hline

Sampling Method & Top-k Sampling \\ \hline
Temperature & 2.0 \\ \hline
Top-k & 40 \\ \hline
Top-p (Nucleus Sampling) & 0.85 \\ \hline
Repetition Penalty & 1.5 \\ \hline

\end{tabular}
\caption{Paraphrasing}
\label{table:dbcv-hyperparams}
\end{table}

\subsection{Hyperparameters and Infrastructure}
\label{Methodological details}

For BERT, computational experiments were run on a machine containing RTX 2080 Ti GPU.
For Llama, computational experiment run on NVIDIA A40 with 2 x AMD EPYC 7413 24-Core Processor. Below are the hyperparameters used for each six methods.

Originally, Llama-3.1-8B-Instruct is a causal language model designed for text generation,  to adapt it for classification, we add a fully trainable classification head on top of the base model. The fully trainable classification head is optimized in conjunction with low-rank adaptation (LoRA) \cite{hu2021lora} adapters. To lower the task's computational cost, we utilize 4-bit quantization. 

\begin{table}[h!]
\centering
\scalebox{1}{ 
\begin{tabular}{|l|l|}
\hline
\textbf{Hyperparameter}  & \textbf{Value}        \\ \hline
Batch size      & 16                   \\ \hline
Learning rate   & 1e-5                 \\ \hline
Max sequence length & 128              \\ \hline
Epochs         & 10                   \\ \hline
Optimizer     & AdamW                \\ \hline
Pooling strategy    & Mean               \\ \hline

\end{tabular}
}
\caption{Hyperparameters for DBV, SBV-ind, SBV-all }
\label{table:dbcv-hyperparams-1}
\end{table}

\begin{table}[h!]
\centering
\scalebox{1}{ 
\begin{tabular}{|l|l|}
\hline
\textbf{Hyperparameter}      & \textbf{Value}             \\ \hline
Batch size          & 16                         \\ \hline
Learning rate       & 1e-5                       \\ \hline
Epochs           & 5                          \\ \hline
Optimizer           & AdamW                      \\ \hline
Pooling strategy    & Mean                       \\ \hline
\end{tabular}
}
\caption{Hyperparameters for DBS  }
\label{table:DBCS}
\end{table}

\begin{table}[h!]
\centering
\scalebox{1}{ 
\begin{tabular}{|l|l|}
\hline
\textbf{Hyperparameter}    & \textbf{Value}             \\ \hline
Batch size        & 16                         \\ \hline
Learning rate       & 1e-5                       \\ \hline
Epochs              & 5                          \\ \hline
Optimizer          & AdamW                      \\ \hline
Margin              & 1.0                        \\ \hline
Alpha (weight for triplet loss) & 1.0            \\ \hline
Pooling strategy        & Mean                   \\ \hline

\end{tabular}
}
\caption{Hyperparameters for DBS-SC}
\label{table:triplet}
\end{table}

\begin{table}[h!]
\centering
\scalebox{1}{ 
\begin{tabular}{|l|l|}
\hline
\textbf{Hyperparameter}           & \textbf{Value}             \\ \hline
Batch size               & 64                         \\ \hline
Learning rate & 1e-5                   \\ \hline
Epochs                  & 5                          \\ \hline
Optimizer          & AdamW                      \\ \hline
Alpha (weight for Contrastive loss) & 5.0         \\ \hline
Pooling strategy         & Mean                       \\ \hline

\end{tabular}
}
\caption{Hyperparameters for DBS-UC }
\label{table:contrastive-tensions}
\end{table}

\begin{table}[h!]
\centering
\scalebox{1}{ 
\begin{tabular}{|l|l|}
\hline
\textbf{Hyperparameter}           & \textbf{Value}             \\ \hline
Batch size               & 8                         \\ \hline
Learning rate & 2e-5                   \\ \hline
Epochs                  & 5, 1                        \\ \hline
Optimizer          & AdamW                      \\ \hline
lora alpha & 16 \\ \hline 
lora dropout  & 0.1 \\ \hline
\end{tabular}
}
\caption{Hyperparameters for IS-single (Llama), and DS-simple (Llama) with 1 epoch }
\label{table:contrastive-tensions-1}
\end{table}


\section{Extended Results on value prediction and subjectivity prediction with STD}

\label{Results on value prediction}

\begin{table*}[tp]
\centering
\resizebox{\textwidth}{!}{%
\begin{tabular}{lcccccccccccccccc}
\toprule
\textbf{Method} & \multicolumn{3}{c}{\textbf{Label 0}} & \multicolumn{3}{c}{\textbf{Label 1}} & \multicolumn{3}{c}{\textbf{Label 2}} & \multicolumn{3}{c}{\textbf{Label 3}} \\

 & \textbf{P} & \textbf{R} & \textbf{F$_{1}$}  & \textbf{P} & \textbf{R} & \textbf{F$_{1}$}  & \textbf{P} & \textbf{R} & \textbf{F$_{1}$} &  \textbf{P} & \textbf{R} & \textbf{F$_{1}$} \\
 
\midrule

DBCV-W015 & 0.54$\pm$0.02 & 0.51$\pm$0.02 & 0.52$\pm$0.02  
   & 0.59$\pm$0.04 & 0.37$\pm$0.02 & 0.45$\pm$0.03 
   & 0.57$\pm$0.04 & 0.37$\pm$0.10 & 0.44$\pm$0.08  
   & 0.87$\pm$0.09 & 0.23$\pm$0.06 & 0.36$\pm$0.07  \\

DBCV-W023 & 0.65$\pm$0.01 & 0.41$\pm$0.11 & 0.49$\pm$0.08 
   & 0.70$\pm$0.05 & 0.65$\pm$0.04 & 0.67$\pm$0.01 
   & 0.64$\pm$0.03 & 0.63$\pm$0.05 & 0.64$\pm$0.02 
   & 0.60$\pm$0.05 & 0.26$\pm$0.10 & 0.36$\pm$0.09 \\
   
DBCV-W018 & 0.67$\pm$0.04 & 0.48$\pm$0.04 & 0.56$\pm$0.02 
   & 0.69$\pm$0.02 & 0.69$\pm$0.04 & 0.69$\pm$0.02  
   & 0.62$\pm$0.03 & 0.67$\pm$0.05 & 0.64$\pm$0.02 
   & 0.63$\pm$0.09 & 0.27$\pm$0.11 & 0.36$\pm$0.08  \\

DBCV-W024 & 0.68$\pm$0.02 & 0.74$\pm$0.03 & 0.71$\pm$0.01 
   & 0.69$\pm$0.05 & 0.62$\pm$0.05 & 0.65$\pm$0.03 
   & 0.66$\pm$0.02 & 0.72$\pm$0.02 & 0.69$\pm$0.01 
   & 0.59$\pm$0.01 & 0.50$\pm$0.09 & 0.54$\pm$0.05 \\

SBDCV-W015 & 0.52$\pm$0.03 & 0.63$\pm$0.09 & 0.57$\pm$0.03 
   & 0.48$\pm$0.12 & 0.47$\pm$0.16 & 0.45$\pm$0.04  
   & 0.37$\pm$0.08 & 0.66$\pm$0.13 & 0.46$\pm$0.05  
   & 0.42$\pm$0.10 & 0.38$\pm$0.15 & 0.38$\pm$0.10  \\

SBDCV-W023 & 0.59$\pm$0.05 & 0.57$\pm$0.11 & 0.57$\pm$0.04 
   & 0.66$\pm$0.05 & 0.66$\pm$0.04 & 0.66$\pm$0.02 
   & 0.60$\pm$0.08 & 0.81$\pm$0.05 & 0.69$\pm$0.05  
   & 0.47$\pm$0.03 & 0.42$\pm$0.06 & 0.44$\pm$0.03  \\

SBDCV-W018 & 0.56$\pm$0.04 & 0.66$\pm$0.03 & 0.60$\pm$0.02  
   & 0.72$\pm$0.04 & 0.63$\pm$0.05 & 0.67$\pm$0.01 
   & 0.53$\pm$0.06 & 0.77$\pm$0.04 & 0.62$\pm$0.03 
   & 0.42$\pm$0.06 & 0.38$\pm$0.13 & 0.38$\pm$0.05  \\

SBDCV-W024 & 0.70$\pm$0.02 & 0.71$\pm$0.04 & 0.71$\pm$0.01 
   & 0.70$\pm$0.03 & 0.61$\pm$0.03 & 0.65$\pm$0.01 
   & 0.67$\pm$0.02 & 0.71$\pm$0.01 & 0.69$\pm$0.01 
   & 0.60$\pm$0.02 & 0.50$\pm$0.06 & 0.54$\pm$0.03  \\

SBCV-W015 & 0.54$\pm$0.01 & 0.62$\pm$0.04 & 0.58$\pm$0.02 
   & 0.63$\pm$0.06 & 0.32$\pm$0.08 & 0.42$\pm$0.07 
   & 0.63$\pm$0.06 & 0.39$\pm$0.07 & 0.48$\pm$0.07 
   & 0.67$\pm$0.15 & 0.35$\pm$0.12 & 0.43$\pm$0.08  \\

SBCV-W023 & 0.63$\pm$0.03 & 0.56$\pm$0.06 & 0.59$\pm$0.03 
   & 0.74$\pm$0.03 & 0.58$\pm$0.08 & 0.64$\pm$0.05  
   & 0.71$\pm$0.04 & 0.65$\pm$0.04 & 0.67$\pm$0.03 
   & 0.51$\pm$0.04 & 0.45$\pm$0.05 & 0.47$\pm$0.02  \\

SBCV-W018 & 0.66$\pm$0.03 & 0.54$\pm$0.05 & 0.59$\pm$0.02  
   & 0.69$\pm$0.03 & 0.71$\pm$0.03 & 0.70$\pm$0.02 
   & 0.68$\pm$0.03 & 0.67$\pm$0.02 & 0.67$\pm$0.02 
   & 0.50$\pm$0.03 & 0.35$\pm$0.05 & 0.41$\pm$0.04  \\

SBCV-W024 & 0.70$\pm$0.03 & 0.75$\pm$0.03 & 0.72$\pm$0.01 
   & 0.69$\pm$0.04 & 0.62$\pm$0.05 & 0.65$\pm$0.02 
   & 0.66$\pm$0.02 & 0.72$\pm$0.03 & 0.69$\pm$0.01  
   & 0.56$\pm$0.02 & 0.59$\pm$0.07 & 0.57$\pm$0.03  \\

SBV-all-Llama-3.1-FT-W015 & 0.68 & 0.43 & 0.53 
   & 0.53 & 0.58 & 0.55  
   & 0.44 & 0.42 & 0.43 
   & 0.67 & 0.12 & 0.21  \\

SBV-all-Llama-3.1-FT-W023 & 0.55 & 0.53 & 0.54  
   & 0.66 & 0.67 & 0.67  
   & 0.65 & 0.77 & 0.71  
   & 0.60 & 0.35 & 0.44  \\

SBV-all-Llama-3.1-FT-W018 & 0.61 & 0.48 & 0.54  
   & 0.70 & 0.81 & 0.75 
   & 0.66 & 0.68 & 0.67  
   & 0.61 & 0.34 & 0.44  \\ 

SBV-all-Llama-3.1-FT-W024 & 0.69 & 0.70 & 0.69  
   & 0.71 & 0.71 & 0.71  
   & 0.70 & 0.56 & 0.73 
   & 0.67 & 0.43 & 0.52  \\

\midrule
\midrule
 & \multicolumn{3}{c}{\textbf{Label 4}} & \multicolumn{3}{c}{\textbf{Label 5}} & \multicolumn{3}{c}{\textbf{Label 6}} & \multicolumn{3}{c}{\textbf{Label 7}} \\

 & \textbf{P} & \textbf{R} & \textbf{F$_{1}$}  & \textbf{P} & \textbf{R} & \textbf{F$_{1}$}  & \textbf{P} & \textbf{R} & \textbf{F$_{1}$} & \textbf{P} & \textbf{R} & \textbf{F$_{1}$}  \\
\midrule

DBCV-W015 & 0.49$\pm$0.14 & 0.28$\pm$0.05 & 0.35$\pm$0.06 
   & 0.63$\pm$0.06 & 0.28$\pm$0.06 & 0.39$\pm$0.05  
   & 0.68$\pm$0.07 & 0.45$\pm$0.04 & 0.54$\pm$0.03 
   & 0.00$\pm$0.00 & 0.00$\pm$0.00 & 0.00$\pm$0.00  \\

DBCV-W023 & 0.56$\pm$0.05 & 0.46$\pm$0.05 & 0.50$\pm$0.04 
   & 0.65$\pm$0.04 & 0.30$\pm$0.08 & 0.41$\pm$0.07 
   & 0.61$\pm$0.03 & 0.43$\pm$0.03 & 0.50$\pm$0.03 
   & 0.77$\pm$0.09 & 0.16$\pm$0.02 & 0.26$\pm$0.03  \\
   
DBCV-W018 & 0.64$\pm$0.02 & 0.48$\pm$0.06 & 0.54$\pm$0.04  
   & 0.75$\pm$0.11 & 0.31$\pm$0.06 & 0.43$\pm$0.04 
   & 0.68$\pm$0.04 & 0.46$\pm$0.06 & 0.55$\pm$0.03 
   & 0.75$\pm$0.03 & 0.27$\pm$0.02 & 0.39$\pm$0.03 \\

DBCV-W024 & 0.61$\pm$0.02 & 0.54$\pm$0.05 & 0.57$\pm$0.02 
   & 0.68$\pm$0.05 & 0.51$\pm$0.07 & 0.58$\pm$0.03 
   & 0.70$\pm$0.04 & 0.49$\pm$0.05 & 0.57$\pm$0.02 
   & 0.58$\pm$0.10 & 0.22$\pm$0.08 & 0.31$\pm$0.07 \\

SBDCV-W015 & 0.33$\pm$0.04 & 0.60$\pm$0.16 & 0.41$\pm$0.03  
   & 0.58$\pm$0.05 & 0.48$\pm$0.10 & 0.52$\pm$0.06 
   & 0.57$\pm$0.10 & 0.67$\pm$0.18 & 0.59$\pm$0.07 
   & 0.06$\pm$0.08 & 0.10$\pm$0.14 & 0.07$\pm$0.10  \\

SBDCV-W023 & 0.50$\pm$0.06 & 0.56$\pm$0.05 & 0.52$\pm$0.02 
   & 0.51$\pm$0.05 & 0.48$\pm$0.08 & 0.49$\pm$0.02 
   & 0.52$\pm$0.05 & 0.54$\pm$0.07 & 0.53$\pm$0.03 
   & 0.56$\pm$0.19 & 0.32$\pm$0.13 & 0.37$\pm$0.03  \\

SBDCV-W018 & 0.53$\pm$0.09 & 0.63$\pm$0.04 & 0.57$\pm$0.03  
   & 0.54$\pm$0.04 & 0.53$\pm$0.06 & 0.53$\pm$0.02  
   & 0.55$\pm$0.06 & 0.57$\pm$0.12 & 0.55$\pm$0.04  
   & 0.58$\pm$0.27 & 0.35$\pm$0.12 & 0.38$\pm$0.03  \\

SBDCV-W024 & 0.61$\pm$0.02 & 0.52$\pm$0.03 & 0.56$\pm$0.02  
   & 0.68$\pm$0.03 & 0.48$\pm$0.05 & 0.56$\pm$0.03 
   & 0.67$\pm$0.02 & 0.50$\pm$0.05 & 0.57$\pm$0.03 
   & 0.65$\pm$0.09 & 0.21$\pm$0.03 & 0.31$\pm$0.02  \\

SBCV-W015 & 0.61$\pm$0.06 & 0.38$\pm$0.06 & 0.47$\pm$0.06 
   & 0.64$\pm$0.11 & 0.48$\pm$0.10 & 0.54$\pm$0.05  
   & 0.74$\pm$0.03 & 0.50$\pm$0.05 & 0.60$\pm$0.04 
   & 0.59$\pm$0.13 & 0.24$\pm$0.06 & 0.34$\pm$0.08  \\

SBCV-W023 & 0.64$\pm$0.03 & 0.43$\pm$0.05 & 0.51$\pm$0.04  
   & 0.52$\pm$0.04 & 0.39$\pm$0.07 & 0.44$\pm$0.04  
   & 0.68$\pm$0.03 & 0.44$\pm$0.07 & 0.53$\pm$0.05  
   & 0.62$\pm$0.08 & 0.31$\pm$0.03 & 0.41$\pm$0.01  \\

SBCV-W018 & 0.67$\pm$0.03 & 0.50$\pm$0.07 & 0.57$\pm$0.04  
   & 0.68$\pm$0.05 & 0.33$\pm$0.09 & 0.44$\pm$0.07 
   & 0.68$\pm$0.05 & 0.48$\pm$0.08 & 0.56$\pm$0.05  
   & 0.67$\pm$0.03 & 0.33$\pm$0.02 & 0.44$\pm$0.02  \\ 

SBCV-W024 & 0.61$\pm$0.02 & 0.55$\pm$0.01 & 0.58$\pm$0.01  
   & 0.65$\pm$0.06 & 0.54$\pm$0.06 & 0.59$\pm$0.02  
   & 0.69$\pm$0.03 & 0.58$\pm$0.05 & 0.63$\pm$0.02 
   & 0.46$\pm$0.03 & 0.42$\pm$0.05 & 0.44$\pm$0.02  \\

SBV-all-Llama-3.1-FT-W015 & 0.55 & 0.32 & 0.41
   & 0.74 & 0.36 & 0.48 
   & 0.88 & 0.54 & 0.67
   & 0.38 & 0.12 & 0.18  \\

SBV-all-Llama-3.1-FT-W023 & 0.77 & 0.40 & 0.52  
   & 0.62 & 0.35 & 0.45
   & 0.73 & 0.47 & 0.57 
   & 0.58 & 0.38 & 0.46 \\

SBV-all-Llama-3.1-FT-W018 & 0.79 & 0.34 & 0.48  
   & 0.86 & 0.30 & 0.44
   & 0.81 & 0.55 & 0.65 
   & 0.77 & 0.25 & 0.38 \\ 

SBV-all-Llama-3.1-FT-W024 & 0.77 & 0.44 & 0.56  
   & 0.73 & 0.45 & 0.56  
   & 0.81 & 0.51 & 0.63 
   & 0.51 & 0.29 & 0.37  \\

\bottomrule
\end{tabular}%
}
\caption{The average and standard deviation of precision, recall, and F-score of value predictions.}
\label{table:Value results}
\end{table*}

\begin{table*}[tp]
\centering
\small
\resizebox{\textwidth}{!}{%
\begin{tabular}{l|ccc|ccc|ccc|ccc}
\toprule
\textbf{Method} & \multicolumn{3}{c|}{\textbf{Achievement}} & \multicolumn{3}{c|}{\textbf{Power: resources}} & \multicolumn{3}{c|}{\textbf{Security: personal}} & \multicolumn{3}{c}{\textbf{Security: societal}} \\

 & \textbf{P} & \textbf{R} & \textbf{F$_{1}$}  & \textbf{P} & \textbf{R} & \textbf{F$_{1}$}  & \textbf{P} & \textbf{R} & \textbf{F$_{1}$} & \textbf{P} & \textbf{R} & \textbf{F$_{1}$}  \\
 
\midrule

IS-each (BERT) & 0.40$\pm$0.01 & 0.56$\pm$0.05 & 0.46$\pm$0.01  
   & 0.41$\pm$0.03 & 0.50$\pm$0.04 & 0.45$\pm$0.03  
   & 0.35$\pm$0.01 & 0.44$\pm$0.03 & 0.39$\pm$0.02  
   & 0.36$\pm$0.01 & 0.50$\pm$0.07 & 0.42$\pm$0.03  \\

IS-shared (BERT) & 0.37$\pm$0.02 & 0.29$\pm$0.05 & 0.32$\pm$0.04  
   & 0.38$\pm$0.02 & 0.26$\pm$0.06 & 0.31$\pm$0.04 
   & 0.32$\pm$0.08 & 0.20$\pm$0.08 & 0.24$\pm$0.08  
   & 0.41$\pm$0.03 & 0.34$\pm$0.05 & 0.37$\pm$0.04  \\

IS-single (BERT) & 0.38$\pm$0.05 & 0.39$\pm$0.05 & 0.38$\pm$0.04  
   & 0.40$\pm$0.01 & 0.33$\pm$0.03 & 0.36$\pm$0.02 
   & 0.37$\pm$0.02 & 0.40$\pm$0.06 & 0.38$\pm$0.03  
   & 0.38$\pm$0.03 & 0.36$\pm$0.06 & 0.37$\pm$0.02  \\

IS-single (Llama) & 0.41 & 0.39  & 0.40
    & 0.44 & 0.29  & 0.35 
    & 0.46 & 0.41  & 0.33
    & 0.45 & 0.31  & 0.37 \\

\midrule

DS-simple (BERT) & 0.81$\pm$0.04 & 0.57$\pm$0.06 & 0.67$\pm$0.03   
      & 0.82$\pm$0.05 & 0.61$\pm$0.07 & 0.69$\pm$0.02   
      & 0.83$\pm$0.03 & 0.66$\pm$0.05 & 0.73$\pm$0.03  
      & 0.89$\pm$0.05 & 0.58$\pm$0.08 & 0.70$\pm$0.04  \\

DS-sup (BERT) & 0.78$\pm$0.03 & 0.62$\pm$0.04 & 0.69$\pm$0.01 
    & 0.84$\pm$0.03 & 0.58$\pm$0.04 & 0.69$\pm$0.02  
    & 0.83$\pm$0.04 & 0.64$\pm$0.06 & 0.72$\pm$0.03 
    & 0.84$\pm$0.04 & 0.65$\pm$0.06 & 0.73$\pm$0.02  \\

DS-unsup (BERT) & 0.65$\pm$0.05  & 0.60 $\pm$0.03 & 0.62$\pm$0.03 
    & 0.73$\pm$0.05 & 0.55$\pm$0.04 & 0.63 $\pm$0.02 
    & 0.65 $\pm$0.06 & 0.72$\pm$0.05 & 0.68$\pm$0.04 
    & 0.71$\pm$0.06 & 0.67$\pm$0.06   & 0.68$\pm$0.02  \\

DS-simple (Llama)  & 0.71 & 0.76 & 0.74
    & 0.94 & 0.65 & 0.77
    & 0.73 & 0.80 & 0.76 
    & 0.82 & 0.74 & 0.78
      \\

\midrule

Baseline (random) & 0.55$\pm$0.01 & 0.49 $\pm$0.02 & 0.51$\pm$0.01 
         & 0.53$\pm$0.01 & 0.50 $\pm$0.01 & 0.52$\pm$0.01 
        & 0.51 $\pm$0.01 & 0.48$\pm$0.01 & 0.49$\pm$0.01 
        & 0.55$\pm$0.01 & 0.49$\pm$0.03 &  0.52$\pm$0.01 \\

\bottomrule
\toprule

 & \multicolumn{3}{c|}{\textbf{Conformity: 
rules}} & \multicolumn{3}{c|}{\textbf{Benevolence: caring}} & \multicolumn{3}{c|}{\textbf{Benevolence: dependability}} & \multicolumn{3}{c}{\textbf{Universalism: concern}} \\

 & \textbf{P} & \textbf{R} & \textbf{F$_{1}$}  & \textbf{P} & \textbf{R} & \textbf{F$_{1}$}  & \textbf{P} & \textbf{R} & \textbf{F$_{1}$} & \textbf{P} & \textbf{R} & \textbf{F$_{1}$}  \\
 
\midrule

IS-each (BERT) & 0.30$\pm$0.03 & 0.35$\pm$0.07 & 0.33$\pm$0.04 
   & 0.33$\pm$0.01 & 0.42$\pm$0.04 & 0.37$\pm$0.01 
   & 0.20$\pm$0.03 & 0.24$\pm$0.00 & 0.22$\pm$0.02 
   & 0.36$\pm$0.06 & 0.2$\pm$0.06 & 0.24$\pm$0.04  \\

IS-shared (BERT) & 0.28$\pm$0.03 & 0.18$\pm$0.06 & 0.22$\pm$0.04 
   & 0.28$\pm$0.02 & 0.18$\pm$0.03 & 0.22$\pm$0.03 
   & 0.14$\pm$0.03 & 0.15$\pm$0.06 & 0.14$\pm$0.04 
   & 0.32$\pm$0.04 & 0.16$\pm$0.10 & 0.20$\pm$0.08  \\

IS-single (BERT) & 0.30$\pm$0.05 & 0.26$\pm$0.04 & 0.27$\pm$0.04 
   & 0.35$\pm$0.03 & 0.27$\pm$0.06 & 0.30$\pm$0.04 
   & 0.23$\pm$0.02 & 0.19$\pm$0.02 & 0.21$\pm$0.02  
   & 0.37$\pm$0.02 & 0.20$\pm$0.05 & 0.25$\pm$0.04  \\

IS-single (Llama)  & 0.42 & 0.19  & 0.26 
    & 0.35 & 0.22  & 0.27   
    & 0.33 & 0.13   & 0.18
    & 0.35 & 0.17 & 0.23  \\

\midrule

DS-simple (BERT) & 0.89$\pm$0.04 & 0.60$\pm$0.06 & 0.71$\pm$0.04 
      & 0.86$\pm$0.06 & 0.62$\pm$0.09 & 0.71$\pm$0.03 
      & 0.90$\pm$0.03 & 0.60$\pm$0.09 & 0.72$\pm$0.05 
      & 0.85$\pm$0.02 & 0.59$\pm$0.04 & 0.70$\pm$0.02  \\

DS-sup (BERT) & 0.89$\pm$0.01 & 0.61$\pm$0.03 & 0.72$\pm$0.02  
    & 0.89$\pm$0.04 & 0.56$\pm$0.08 & 0.69$\pm$0.04 
    & 0.86$\pm$0.03 & 0.67$\pm$0.06 & 0.75$\pm$0.03 
    & 0.85$\pm$0.04 & 0.59$\pm$0.07 & 0.69$\pm$0.04 \\

DS-unsup (BERT) & 0.70$\pm$0.07 & 0.68 $\pm$0.06 & 0.69$\pm$0.04 
         & 0.78$\pm$0.05 & 0.52 $\pm$0.05 & 0.62$\pm$0.04 
        & 0.66 $\pm$0.05 & 0.55$\pm$0.06 & 0.60$\pm$0.03 
        & 0.70$\pm$0.06 & 0.59$\pm$0.06 &  0.64$\pm$0.03 \\

DS-simple (Llama) 
    & 0.88 & 0.75 & 0.81
    & 0.88 & 0.78 & 0.83
    & 0.88 & 0.86 & 0.87
    & 0.91 & 0.71 & 0.80
      \\

\midrule

Baseline (random) & 0.50$\pm$0.02  & 0.50 $\pm$0.02 & 0.50$\pm$0.02 
         & 0.52$\pm$0.02  & 0.50 $\pm$0.02 & 0.51$\pm$0.02 
        & 0.52 $\pm$0.01  & 0.52$\pm$0.02 & 0.52$\pm$0.01 
        & 0.53$\pm$0.02  & 0.50$\pm$0.02 &  0.52$\pm$0.01 \\

\bottomrule
\end{tabular}%
}
\caption{Precision (P), recall (R), and $F_1$-score of the subjectivity prediction per value. STD shows the 5 runs on different train and validation sets with a fixed test set.}
\label{table:results-appendix}
\end{table*}

\end{document}